\documentclass{article}
\usepackage{great}

\usepackage[utf8]{inputenc}
\usepackage[T1]{fontenc}
\usepackage[hidelinks]{hyperref}
\usepackage{url}
\usepackage{booktabs}
\usepackage{amsfonts}
\usepackage{nicefrac}
\usepackage{microtype}
\usepackage{xcolor}
\usepackage{bm}
\usepackage{amsmath}
\usepackage{amssymb}
\usepackage{graphicx}
\usepackage{multirow}

\makeatletter
\usepackage{xspace}
\DeclareRobustCommand\onedot{\futurelet\@let@token\@onedot}
\def\@onedot{\ifx\@let@token.\else.\null\fi\xspace}
\def\eg{\emph{e.g}\onedot}

\def\etal{\emph{et al}\onedot}
\def\etc{\emph{etc}\onedot}
\makeatother

\makeatletter
\newcommand*{\centerfloat}{%
	\parindent \z@
	\leftskip \z@ \@plus 1fil \@minus \textwidth
	\rightskip\leftskip
	\parfillskip \z@skip}
\makeatother

\title{View-oriented Conversation Compiler for \\Agent Trace Analysis}

\author{\texttt{Lvmin Zhang and Maneesh Agrawala}\\\texttt{Stanford University}}

\begin{document}

\maketitle

\begin{abstract}
\vspace{10pt}
We observe that an agent trace is a structured document. A coding agent session contains user turns, assistant text, chain of thought blocks, tool calls, tool results, subagent invocations, compaction boundaries, and harness injected directives, and may exceed ten thousand JSONL lines. Effective trace analysis requires a lossless record, a session level overview, and content retrieval with conversational roles. To this end, we propose VCC, namely View oriented Conversation Compiler, which lexes, parses, and lowers a raw JSONL log into three views based on one intermediate representation. The full view provides the lossless transcript and defines the line number coordinate system. The UI view reconstructs the interaction perceived by the user. The adaptive view selects relevant trace content and preserves role annotations and line range pointers. Line numbers are assigned before view lowering, ensuring that all pointers can be resolved to the full view. We evaluate VCC in a context engineering experiment on AppWorld by varying the input format of the reflector. Experiments with three model configurations and two test splits show that VCC improves aggregate task goal pass rates by 1.1 to 4.2 points, reduces reflector token consumption by one half to two thirds, and generates smaller memory files. These results demonstrate the effectiveness of trace format as an important component of context engineering infrastructure.

\end{abstract}

\section{Introduction}
%% ============================================================

Agent traces provide increasing analytical value in many applications. Context learning trains memory from traces~\cite{zhang2025agentic,fang2026trajectory,suzgun2025dynamic,song2024trial}, memory systems retrieve useful information from traces~\cite{packer2023memgpt,sarthi2024raptor}, and tool use methods learn interaction patterns from traces~\cite{schick2023toolformer}. Harness designers also inspect traces to locate agent failures. In these applications, LLMs or human analysts usually consume traces through plain text, JSON, YAML, and grep.

A contemporary agent trace is a complex structured document. One session of a coding agent such as Claude Code~\cite{anthropics2025claude} or Codex~\cite{openai2025codex} may generate a JSONL log with over ten thousand lines. The log contains user turns, assistant text, chain of thought blocks~\cite{wei2022chain}, tool call parameters, tool result bodies, subagent invocations, context compaction boundaries, base64 encoded media, and system directives injected by the harness. Tool results usually occupy most tokens, while important decision points are distributed among file contents and terminal outputs.

Effective trace analysis requires three properties. First, a lossless record preserves exact tool inputs and outputs for verifying agent decisions. Second, a session level overview presents the sequence of user requests, assistant actions, and execution outcomes. Third, role aware content retrieval distinguishes different semantics of the same keyword, such as an intention in a tool call and an observation in a tool result.

To satisfy these requirements, we propose VCC, namely View oriented Conversation Compiler, which transforms a raw JSONL log into three views based on a shared intermediate representation (IR). VCC adopts a compiler pipeline including lexing, parsing into a typed IR, line number assignment, and view specific lowering. The \emph{full view} outputs every IR node verbatim and provides a shared line number coordinate system. The \emph{UI view} summarizes each tool call in one line with a pointer to the full view, elides internal content, and merges consecutive assistant turns. The \emph{adaptive view} projects the trace through a relevance predicate $\rho$ and preserves the structural skeleton, semantic role, and line range pointer of each matched block. Line numbers are assigned once before lowering, allowing every pointer from different views to accurately locate the corresponding content in the full view.

We evaluate VCC through a context engineering experiment on AppWorld~\cite{trivedi2024appworld}. A generator agent executes tasks, while a reflector agent reads the execution trace and updates a shared memory file. The reflector input is configured as raw JSONL or VCC views for comparison. Across three model configurations, including Opus, Sonnet, and Haiku, and two test splits containing 168 and 416 tasks, VCC improves aggregate task goal pass rates, reduces reflector token consumption by one half to two thirds, and generates smaller memory files.

The main contributions of this work are summarized as follows. First, we formulate agent traces as structured documents and study trace presentation as an important variable of context engineering. Second, we propose VCC with three complementary views sharing one line number coordinate system. Third, we systematically evaluate the influence of trace formats on AppWorld by controlling the reflector input.

\section{VCC}
\label{sec:vcc}
%% ============================================================

\subsection{Three Views}
\label{sec:views}

\begin{figure}[p]
    \centering
    \includegraphics[width=\linewidth]{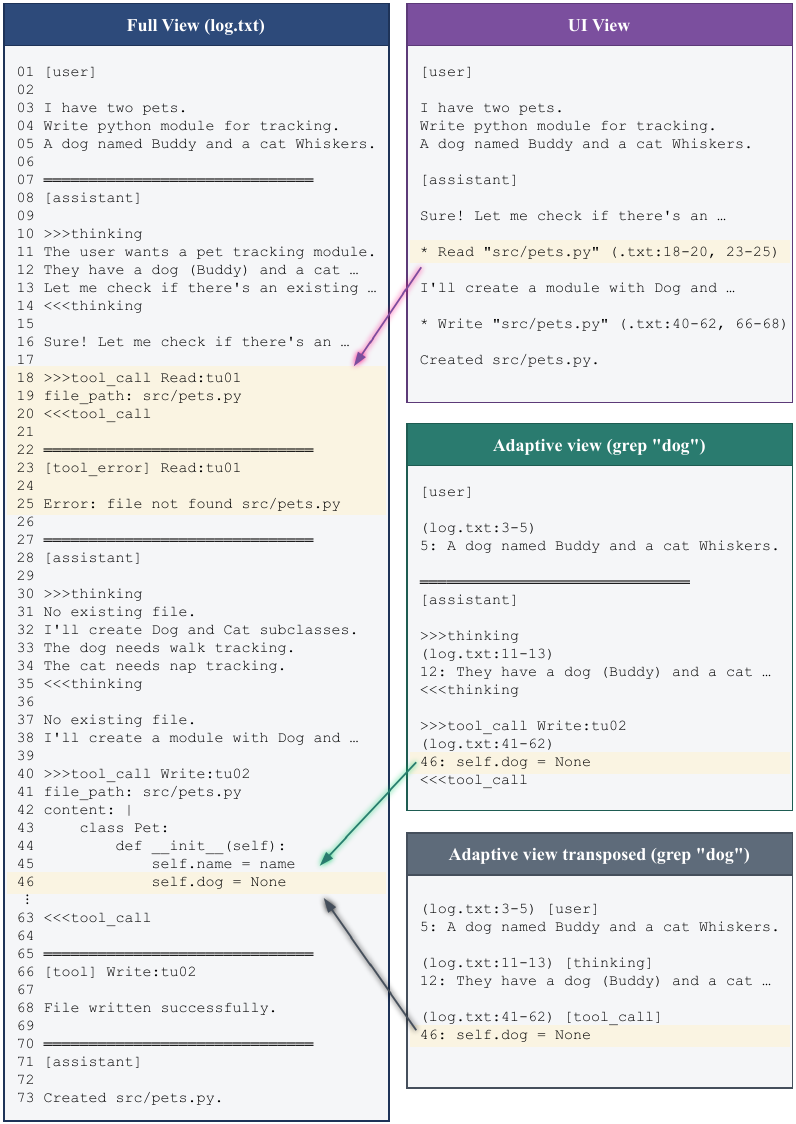}%drawio
    \caption{The same agent trace rendered under three views. The full view (left) emits every IR node with line numbers. The UI view (center) collapses tool calls to one-line summaries with line-range pointers (purple arrows). The adaptive view (right, grep ``dog'') retains only matching blocks with role annotations and range pointers (blue arrows). Highlighted lines show matching content; all pointers resolve to the full view's line-number coordinate system.}
    \label{fig:views}
\end{figure}

VCC models an agent trace as a structured document and provides three views, as shown in Figure~\ref{fig:views}. A trace contains temporally ordered heterogeneous records, including user utterances, assistant responses, chain-of-thought reasoning, tool invocations, tool results, and system directives. The three views are projected from a shared IR with different resolutions and coverage, while using the same line number coordinate system.

The full view is an identity projection. Let $\mathcal{I} = (n_1, n_2, \ldots, n_N)$ denote the IR node sequence after parsing and line assignment. It emits all nodes verbatim:
\begin{equation}
V_{\text{full}} = \bigoplus_{i=1}^{N} \text{emit}(n_i),
\end{equation}
where $\bigoplus$ denotes concatenation with structurally determined blank-line insertion. The rendering is bijective up to serialization whitespace, and provides line number references for the other two views.

The UI view reconstructs the interaction perceived by users. For terminal-based coding agents, it retains assistant text, turn structures, and one-line tool-call summaries. Internal content, including tool-call bodies, tool-result bodies, thinking blocks, and system messages, is elided. Each tool call is represented by a one-line summary with line-range pointers into $V_{\text{full}}$, \eg \texttt{* Read "src/config.py" (file.txt:19-21,24-34)}. Adjacent assistant sections within one user turn are merged into a continuous response. Harness-generated markup, including IDE selection contexts, system reminders, and hook outputs, is filtered to preserve intentional user inputs.

The adaptive view performs structure-preserving projection according to a relevance predicate. Given a predicate $\rho: \text{line} \to \{0, 1\}$, it retains section headers, turn boundaries, and block delimiters, and projects each content block through $\rho$:
\begin{equation}
V_{\text{adapt}}(b) = \begin{cases} \text{match\_lines}(b, \rho) & \text{if } \exists\, l \in [s,e]: \rho(l) = 1 \\ \varnothing & \text{otherwise,} \end{cases}
\end{equation}
where $b$ spans lines $[s, e]$ in $V_{\text{full}}$. The function $\text{match\_lines}$ returns matched lines with their $V_{\text{full}}$ line numbers and a block-level pointer $(f\texttt{:L}s\texttt{-L}e)$. Sections with zero matched content are elided. The predicate $\rho$ supports various implementations, including regex matching, BM25 scoring, embedding similarity, and LLM relevance judgment.

The adaptive view provides two transposed output modalities. The document-oriented modality follows temporal structure and organizes matched blocks by section. The index-oriented modality emits a flat sequence of matched blocks with semantic role tags, \eg \texttt{[tool\_call]}, \texttt{[assistant]}, and \texttt{[thinking]}. Both modalities contain identical data with row-major iteration by section and column-major iteration by match, respectively.

Two related families are multi-level memory and flat search methods. Multi-level memory methods store derived summaries, embeddings, or graph nodes and retrieve information through the stored abstractions. VCC generates query-time projections from one flat IR, where each predicate produces a new projection containing source content and recovery coordinates. Flat search methods organize matched lines as a collection. The adaptive view further preserves semantic roles and block ranges. Therefore, \texttt{[tool\_call] Edit} represents an intention, \texttt{[tool\_result] Read} represents an observation, and the complete context of each block can be recovered through one file pointer.

\subsection{Compiler Pipeline and Line-Number Consistency}
\label{sec:pipeline}
%% ------------------------------------------------------------

VCC adopts a compiler pipeline consisting of four stages. First, lexing tokenizes the JSONL source into typed records and filters records with zero conversational content, including \texttt{queue-operation}, \texttt{file-history-snapshot}, \texttt{progress}, and \texttt{api\_error}. Second, parsing transforms the records into a typed IR. Each node contains a semantic type (\texttt{user}, \texttt{assistant}, \texttt{tool\_call}, \texttt{tool\_result}, \texttt{thinking}, \etc), content lines, and structural metadata such as section and block indices. Third, line assignment sequentially assigns monotonically increasing line numbers to all IR nodes. This deterministic operation is performed once before view lowering. Finally, view lowering projects the shared IR into three views. The full view emits all nodes, while the UI and adaptive views select, truncate, or annotate the corresponding content. All lowering operations preserve the assigned line numbers.

Therefore, pointer consistency across views is a structural invariant of the pipeline. The three views share the same IR and line number assignment, so each pointer can directly locate the corresponding content in the full view. This mechanism is similar to static single assignment form in compilers, where use and definition consistency is guaranteed by the structure.

%% ------------------------------------------------------------
\subsection{Compiler Transformations}
\label{sec:transforms}
%% ------------------------------------------------------------

The lexer and parser process the source language, which is the Claude Code JSONL format in this work. The IR, line assignment, and view lowering modules can be reused for different source languages. Supporting another agent runtime, \eg Codex or OpenClaw, only requires the corresponding lexer and parser.

The Claude Code lexer and parser perform several transformations. Tool call parameters are represented as escaped JSON strings, \eg \texttt{\{"file\_path":"src/pets.py","content":"class Pet:\textbackslash n\quad def \_\_init\_\_..."\}}, and are converted into block-indented text with block scalars (\texttt{|}) for multi-line values. For Read tool results, the \texttt{digits$\to$content} prefix is removed to recover the original source. Harness injected XML markup, including \texttt{<system-reminder>}, \texttt{<ide\_opened\_file>}, and \texttt{<command-message>}, is filtered, and user turns containing only such markup are eliminated. Internal tool calls such as \texttt{TodoWrite} and \texttt{ToolSearch}, ANSI escape codes, and control characters are also removed. Assistant messages split into two JSONL records with the same \texttt{message.id} during context compaction are merged into one section. Records with zero conversational content, including \texttt{queue-operation}, \texttt{file-history-snapshot}, \texttt{progress}, and \texttt{api\_error}, are discarded during lexing. Base64 encoded inline images are decoded into separate files and represented by placeholders.

%% ------------------------------------------------------------
\subsection{Agent Workflow}
\label{sec:workflow}
%% ------------------------------------------------------------

The three views enable a progressive disclosure workflow. An agent or human analyst first reads the UI view to understand the session trajectory, including user requests, assistant actions, and outcomes, while tool I/O and chain of thought details are folded. Then, targeted searches in the adaptive view locate interested content blocks, where each block is annotated with its semantic role and line range pointer. Following the pointer into the full view retrieves the verbatim block content and its surrounding context:
\begin{equation}
V_{\text{ui}} \xrightarrow{\text{pointer}} V_{\text{full}}[s{:}e], \qquad V_{\text{adapt}} \xrightarrow{\text{pointer}} V_{\text{full}}[s{:}e].
\end{equation}
In the experiment of Section~\ref{sec:experiment}, the reflector first reads the UI view to obtain an overall understanding of the session and then follows the pointers to inspect specific tool calls and results in the full view.

\section{Experiment \& Results}
\label{sec:experiment}
%% ============================================================

\subsection{Setup}

We use AppWorld~\cite{trivedi2024appworld}, an API interaction environment for autonomous agents, with an epoch based context engineering protocol. The benchmark contains three splits, including \emph{train} with 90 tasks for learning, \emph{test\_normal} with 168 tasks and \emph{test\_challenge} with 416 tasks for evaluation. All tasks cover three difficulty levels, including d1, d2, and d3. In each epoch, a \emph{generator} agent executes every train task once, and the AppWorld harness evaluates the execution result. Then, a \emph{reflector} agent analyzes the evaluation result and execution trace, and updates a shared memory file (\texttt{MEMORY.md}) with new rules. For failed tasks, the reflector writes \texttt{[CRITICAL]} rules about root causes and preventive measures. For successful tasks, it records risky or suboptimal patterns. The process runs for two epochs (90 tasks $\times$ 2 = 180 analyses per condition). During training, 32 tasks are executed in parallel, and the memory edits from different workers are integrated after each batch by an LLM based diff and merge step. Finally, the learned \texttt{MEMORY.md} is frozen and evaluated on both test splits in one pass.

The investigated variable is the trace format provided to the reflector. In the \emph{VCC} condition, the raw JSONL trace is compiled into different views. The reflector uses the UI view for session level orientation and the full view for line level details, with line number pointers connecting the two views. In the \emph{JSON} condition, the reflector directly reads the raw JSONL, following the default setting of trajectory learning methods. We evaluate three generator and reflector pairings, including Opus+Opus, Sonnet+Sonnet, and Haiku+Sonnet. In our runs, the Haiku reflector was unable to raise pass rates over any candidate. Therefore, the Haiku generator is paired with the Sonnet reflector. For each generator, we further conduct baseline evaluation on both test splits using the initial seed memory with zero trained rules.

We report the official AppWorld metrics, including \emph{task\_goal\_completion}, which measures whether the primary task objective is achieved, and \emph{scenario\_goal\_completion}, which additionally evaluates scenario level constraints about side effects, state consistency, and edge cases. Results are reported for each difficulty level, where d1 = easy, d2 = medium, and d3 = hard. We also report the total reflector token consumption during training (\emph{T tok}), the token cost per analysis (\emph{T/a}), and the character size of the learned memory file (\emph{Mem}).

\subsection{Results}
\label{sec:results}
%% ------------------------------------------------------------

Table~\ref{tab:results} presents the results of different model configurations in terms of pass rate, token consumption, and memory size.

\begin{table}[t]
\centering
\caption{AppWorld results across three model configurations. All pass rates in \%. TN = test\_normal (168 tasks), TC = test\_challenge (416 tasks). T tok = total reflector tokens during training, T/a = tokens per analysis, Mem = learned memory size (characters). Between JSON and VCC, the higher pass rate and the lower cost are in \textbf{bold}.}
\label{tab:results}

\vspace{4pt}
{\small (a) Claude Code 2.1.81: claude-opus-4-6 (generator) + claude-opus-4-6 (reflector)}
\vspace{2pt}

\resizebox{\textwidth}{!}{%
\begin{tabular}{l cccc cccc cccc cccc ccc}
\toprule
& \multicolumn{4}{c}{TN task\_goal} & \multicolumn{4}{c}{TN scenario\_goal} & \multicolumn{4}{c}{TC task\_goal} & \multicolumn{4}{c}{TC scenario\_goal} & & & \\
\cmidrule(lr){2-5} \cmidrule(lr){6-9} \cmidrule(lr){10-13} \cmidrule(lr){14-17}
Method & all & d1 & d2 & d3 & all & d1 & d2 & d3 & all & d1 & d2 & d3 & all & d1 & d2 & d3 & T tok & T/a & Mem \\
\midrule
Baseline & 79.2 & 77.2 & 75.0 & 84.1 & 67.9 & 68.4 & 56.2 & 76.2 & 75.8 & 69.4 & 76.7 & 77.4 & 64.0 & 54.2 & 66.0 & 66.1 & --- & --- & --- \\
JSON & 92.9 & 96.5 & 93.8 & \textbf{88.9} & 87.5 & 89.5 & 87.5 & \textbf{85.7} & 84.7 & 91.7 & \textbf{84.7} & 82.0 & 70.5 & 83.3 & \textbf{74.0} & 63.1 & 22.1M & 124K & 12.6K \\
VCC & \textbf{94.0} & \textbf{98.2} & \textbf{95.8} & \textbf{88.9} & \textbf{91.1} & \textbf{94.7} & \textbf{93.8} & \textbf{85.7} & \textbf{86.3} & \textbf{93.1} & \textbf{84.7} & \textbf{85.1} & \textbf{74.1} & \textbf{87.5} & 70.0 & \textbf{72.3} & \textbf{7.6M} & \textbf{43K} & \textbf{12.0K} \\
\bottomrule
\end{tabular}%
}

\vspace{8pt}
{\small (b) Claude Code 2.1.81: claude-sonnet-4-6 (generator) + claude-sonnet-4-6 (reflector)}
\vspace{2pt}

\resizebox{\textwidth}{!}{%
\begin{tabular}{l cccc cccc cccc cccc ccc}
\toprule
& \multicolumn{4}{c}{TN task\_goal} & \multicolumn{4}{c}{TN scenario\_goal} & \multicolumn{4}{c}{TC task\_goal} & \multicolumn{4}{c}{TC scenario\_goal} & & & \\
\cmidrule(lr){2-5} \cmidrule(lr){6-9} \cmidrule(lr){10-13} \cmidrule(lr){14-17}
Method & all & d1 & d2 & d3 & all & d1 & d2 & d3 & all & d1 & d2 & d3 & all & d1 & d2 & d3 & T tok & T/a & Mem \\
\midrule
Baseline & 73.2 & 77.2 & 58.3 & 81.0 & 67.9 & 73.7 & 50.0 & 76.2 & 61.4 & 61.1 & 59.3 & 63.1 & 50.4 & 45.8 & 50.0 & 52.3 & --- & --- & --- \\
JSON & 90.5 & \textbf{98.2} & 91.7 & 82.5 & 82.1 & \textbf{94.7} & 81.2 & 71.4 & 71.0 & 86.1 & 66.7 & 68.7 & 60.4 & 75.0 & 54.0 & 60.0 & 20.5M & 115K & 23.4K \\
VCC & \textbf{92.9} & 96.5 & \textbf{95.8} & \textbf{87.3} & \textbf{89.3} & \textbf{94.7} & \textbf{93.8} & \textbf{81.0} & \textbf{74.3} & \textbf{88.9} & \textbf{68.7} & \textbf{73.3} & \textbf{63.3} & \textbf{79.2} & \textbf{56.0} & \textbf{63.1} & \textbf{8.8M} & \textbf{49K} & \textbf{15.5K} \\
\bottomrule
\end{tabular}%
}

\vspace{8pt}
{\small (c) Claude Code 2.1.81: claude-haiku-4-5 (generator) + claude-sonnet-4-6 (reflector)}
\vspace{2pt}

\resizebox{\textwidth}{!}{%
\begin{tabular}{l cccc cccc cccc cccc ccc}
\toprule
& \multicolumn{4}{c}{TN task\_goal} & \multicolumn{4}{c}{TN scenario\_goal} & \multicolumn{4}{c}{TC task\_goal} & \multicolumn{4}{c}{TC scenario\_goal} & & & \\
\cmidrule(lr){2-5} \cmidrule(lr){6-9} \cmidrule(lr){10-13} \cmidrule(lr){14-17}
Method & all & d1 & d2 & d3 & all & d1 & d2 & d3 & all & d1 & d2 & d3 & all & d1 & d2 & d3 & T tok & T/a & Mem \\
\midrule
Baseline & 39.9 & 49.1 & 27.1 & 41.3 & 25.0 & 36.8 & 12.5 & 23.8 & 33.8 & 31.9 & 34.0 & 34.4 & 17.3 & 16.7 & 22.0 & 13.9 & --- & --- & --- \\
JSON & 71.4 & 86.0 & \textbf{79.2} & 52.4 & \textbf{62.5} & \textbf{79.0} & \textbf{75.0} & 38.1 & 53.5 & 69.4 & 45.3 & \textbf{53.8} & 36.0 & 45.8 & 30.0 & \textbf{36.9} & 23.3M & 131K & 43.3K \\
VCC & \textbf{75.6} & \textbf{87.7} & 75.0 & \textbf{65.1} & \textbf{62.5} & 68.4 & 56.2 & \textbf{61.9} & \textbf{54.7} & \textbf{81.9} & \textbf{47.3} & 50.3 & \textbf{39.6} & \textbf{66.7} & \textbf{38.0} & 30.8 & \textbf{12.6M} & \textbf{71K} & \textbf{37.9K} \\
\bottomrule
\end{tabular}%
}
\end{table}

As shown in Table~\ref{tab:results}, VCC achieves higher aggregate task\_goal\_completion than JSON on all six model and data split combinations. Specifically, VCC improves the test\_normal and test\_challenge results from 92.9 to 94.0 and from 84.7 to 86.3 for Opus, from 90.5 to 92.9 and from 71.0 to 74.3 for Sonnet, and from 71.4 to 75.6 and from 53.5 to 54.7 for Haiku. Similar improvements are observed on aggregate scenario\_goal\_completion, where VCC performs better on five combinations and obtains the same 62.5 result on Haiku test\_normal. The largest task\_goal improvement appears on Haiku test\_normal, where VCC exceeds JSON by 4.2 points and improves the baseline by 35.7 points. These results indicate that the UI view and pointer navigation help the reflector capture useful decision points, especially for generator traces with more errors.

VCC also substantially reduces reflector token consumption. The total token costs decrease from 22.1M to 7.6M for Opus, from 20.5M to 8.8M for Sonnet, and from 23.3M to 12.6M for Haiku with Sonnet reflector. Correspondingly, the per analysis costs decrease from 124K to 43K, from 115K to 49K, and from 131K to 71K. This demonstrates that the compressed UI view provides an efficient session overview, while pointers support selective access to detailed blocks.

Moreover, VCC produces smaller memory files in all configurations, with 12.0K, 15.5K, and 37.9K characters compared with 12.6K, 23.4K, and 43.3K for JSON. The smaller memory files consistently accompany higher aggregate pass rates, showing that VCC facilitates more focused rules with fewer redundant entries.

JSON obtains higher results on a few difficulty specific cells, such as Haiku d2 on test\_normal task\_goal with 79.2 compared with 75.0. Complete token access may provide additional local details for these cases. Overall, VCC achieves higher pass rates, lower token consumption, and smaller memory files across all configurations. These consistent results show that structured views guide the reflector toward decision relevant information and support precise and concise rule learning.

\section{Related Work}
\label{sec:related}
%% ============================================================

Existing studies investigate the effects of input format, context length, and agent interfaces on LLM performance. He~\etal~\cite{he2024does} show that representing identical content with plain text, Markdown, JSON, or YAML causes performance differences of up to 40 percentage points. Tam~\etal~\cite{tam2024let} find that structured output constraints reduce reasoning ability, while Sui~\etal~\cite{sui2023table} report an accuracy difference of approximately 7 points among table serialization formats. For context length, Liu~\etal~\cite{liu2023lost} observe a U shaped attention pattern where information in the middle of context receives limited attention, and Du~\etal~\cite{du2025context} show that increasing input length causes performance reductions from 13.9\% to 85\% under perfect retrieval. In agent trajectories, Xiao~\etal~\cite{xiao2025reducing} find that redundant tokens account for 40\% to 60\%, and Lindenbauer~\etal~\cite{lindenbauer2025complexity} show that simple observation masking achieves comparable performance with LLM summarization at half the cost. Hong~\etal~\cite{hong2025context} further analyze the performance reduction caused by increasing input length. Liu~\etal~\cite{liu2025beyond} jointly optimize prompt content and format to achieve higher scores than prompt content optimization. Yang~\etal~\cite{yang2024swe} demonstrate that agent computer interface design has an influence comparable to the underlying model, and OpenHands~\cite{wang2024openhands} provides an open platform for interface evaluation. Mei~\etal~\cite{mei2025survey} provide a comprehensive survey of context engineering.

Another line of works extracts reusable experience from agent trajectories. ReAct~\cite{yao2022react} interleaves reasoning traces and actions, Reflexion~\cite{shinn2023reflexion} stores verbal self reflections in episodic memory, and ExpeL~\cite{zhao2023expel} compares successful and failed trajectories to extract general rules. TIM~\cite{fang2026trajectory} derives strategy and recovery tips from trajectories and improves AppWorld by 14.3 percentage points. ACE~\cite{zhang2025agentic} evolves a playbook through generator, reflector, and curator roles, while Dynamic Cheatsheet~\cite{suzgun2025dynamic} maintains a reusable strategy table during testing. ETO~\cite{song2024trial} learns from contrastive trajectory pairs through DPO, ECHO~\cite{hu2025sample} generates counterfactual successful trajectories through hindsight rewriting, and RPMS~\cite{yuan2026rpms} performs conflict arbitration between rules and memory. AutoGuide~\cite{fu2024autoguide} generates conditional guidelines from offline experience, AutoManual~\cite{chen2024automanual} constructs instruction manuals through multiple agent interaction, and Agent Workflow Memory~\cite{wang2024agent} induces reusable workflows with a 51.1\% improvement on WebArena. MACLA~\cite{forouzandeh2025learning} further distills trajectories into procedural memory with Bayesian reliability tracking.

Several works model agent traces as structured objects. AgentRx~\cite{barke2026agentrx} normalizes heterogeneous logs into a common intermediate representation and synthesizes executable constraints, improving failure localization by 23.6 percentage points. TraceSIR~\cite{yang2026tracesir} employs a multiple agent framework with an LLM importance scorer for trace reduction. The Cognition Agent Trace specification~\cite{cognition0000cognition} defines a structured JSON format for agent provenance, and AgentStepper~\cite{hutter2026agentstepper} represents trajectories as structured conversations with interactive breakpoints and stepwise execution. TRAIL~\cite{deshpande2025trail} evaluates LLM trace debugging on 148 human annotated traces and reports a best score of 11\%. AgenTracer~\cite{zhang2025agentracer} reports failure attribution accuracy lower than 10\% for current reasoning LLMs on multiple agent traces. DoVer~\cite{ma2025dover} combines hypothesis based debugging with active intervention and replay, recovering 18\% to 28\% of failed trials. LOCA-bench~\cite{zeng2026loca} evaluates agent performance under controllable context growth. DSPy~\cite{khattab2023dspy} compiles prompt pipeline parameters, while SGLang~\cite{zheng2023sglang} compiles LLM programs for efficient execution.

Agent memory and context management have also attracted increasing attention. MemGPT~\cite{packer2023memgpt}, A-MEM~\cite{xu2025mem}, Mem0~\cite{chhikara2025mem}, and RAPTOR~\cite{sarthi2024raptor} organize stored experience with hierarchical or graph structures, following the memory stream architecture of Park~\etal~\cite{park2023generative}. Hu~\etal~\cite{hu2025memory} survey the broader development of agent memory. Sumers~\etal~\cite{sumers2023cognitive} propose CoALA, which organizes agents with modular memory and structured action spaces. Context compression methods include LLMLingua~\cite{jiang2023llmlingua}, which achieves up to 20 times prompt compression, AgentFold~\cite{ye2025agentfold} and ACON~\cite{kang2025acon} for runtime observation compression, and AgentRM~\cite{she2026agentrm} and AgentOCR~\cite{feng2026agentocr} for learned compression. Focus~\cite{verma2026active} enables agents to determine when to consolidate and prune history, while Context-Folding~\cite{sun2025scaling} folds completed subtasks and reduces active context by 10 times. Multiple agent frameworks such as AutoGen~\cite{wu2023autogen} generate traces with nested structures. LangSmith~\cite{langchainai2023langsmith}, Langfuse~\cite{langfuse2023langfuse}, and AgentTrace~\cite{alsayyad2026agenttrace} provide structured telemetry for running agents. Memory pointer approaches~\cite{labate2025solving} manage context window overflow through pointers to stored content.

\section{Conclusion}
%% ============================================================

In this paper, we present VCC with three complementary views and study trace format as a variable of context engineering. VCC compiles raw JSONL logs into views with consistent line numbers and pointer navigation. With the reflector input format as the controlled variable, experiments on AppWorld show that VCC improves pass rates, reduces token consumption, and decreases learned memory across three model configurations. These results demonstrate that agent traces are structured documents and their presentation plays an important role in context engineering infrastructure. In future work, we will evaluate VCC on SWE-bench~\cite{jimenez2023swe}, TAU-bench~\cite{yao2024tau}, WebArena~\cite{zhou2023webarena}, AgentBench~\cite{liu2023agentbench}, and additional trajectory learning methods.

\bibliographystyle{abbrvnat}
\bibliography{refs,extras}

\begin{thebibliography}{63}
\providecommand{\natexlab}[1]{#1}
\providecommand{\url}[1]{\texttt{#1}}
\expandafter\ifx\csname urlstyle\endcsname\relax
  \providecommand{\doi}[1]{doi: #1}\else
  \providecommand{\doi}{doi: \begingroup \urlstyle{rm}\Url}\fi

\bibitem[AlSayyad et~al.(2026)AlSayyad, Huang, and Pal]{alsayyad2026agenttrace}
A.~AlSayyad, K.~Y. Huang, and R.~Pal.
\newblock Agenttrace: A structured logging framework for agent system observability, 2026.
\newblock URL \url{https://arxiv.org/abs/2602.10133}.

\bibitem[{anthropics}(2025)]{anthropics2025claude}
{anthropics}.
\newblock claude-code.
\newblock \url{https://github.com/anthropics/claude-code}, 2025.
\newblock URL \url{https://github.com/anthropics/claude-code}.
\newblock Claude Code is an agentic coding tool that lives in your terminal, understands your codebase, and helps you code faster by executing routine tasks, explaining complex code, and handling git workflows - all through natural language commands.

\bibitem[Barke et~al.(2026)Barke, Goyal, Khare, Singh, Nath, and Bansal]{barke2026agentrx}
S.~Barke, A.~Goyal, A.~Khare, A.~Singh, S.~Nath, and C.~Bansal.
\newblock Agentrx: Diagnosing ai agent failures from execution trajectories, 2026.
\newblock URL \url{https://arxiv.org/abs/2602.02475}.

\bibitem[Chen et~al.(2024)Chen, Li, Yang, Yu, Lin, and He]{chen2024automanual}
M.~Chen, Y.~Li, Y.~Yang, S.~Yu, B.~Lin, and X.~He.
\newblock Automanual: Constructing instruction manuals by llm agents via interactive environmental learning, 2024.
\newblock URL \url{https://arxiv.org/abs/2405.16247}.

\bibitem[Chhikara et~al.(2025)Chhikara, Khant, Aryan, Singh, and Yadav]{chhikara2025mem}
P.~Chhikara, D.~Khant, S.~Aryan, T.~Singh, and D.~Yadav.
\newblock Mem0: Building production-ready ai agents with scalable long-term memory, 2025.
\newblock URL \url{https://arxiv.org/abs/2504.19413}.

\bibitem[{Cognition}()]{cognition0000cognition}
{Cognition}.
\newblock Cognition.
\newblock Cognition.
\newblock URL \url{https://cognition.ai/blog/agent-trace}.

\bibitem[Deshpande et~al.(2025)Deshpande, Gangal, Mehta, Krishnan, Kannappan, and Qian]{deshpande2025trail}
D.~Deshpande, V.~Gangal, H.~Mehta, J.~Krishnan, A.~Kannappan, and R.~Qian.
\newblock Trail: Trace reasoning and agentic issue localization, 2025.
\newblock URL \url{https://arxiv.org/abs/2505.08638}.

\bibitem[Du et~al.(2025)Du, Tian, Ronanki, Rongali, Bodapati, Galstyan, Wells, Schwartz, Huerta, and Peng]{du2025context}
Y.~Du, M.~Tian, S.~Ronanki, S.~Rongali, S.~Bodapati, A.~Galstyan, A.~Wells, R.~Schwartz, E.~A. Huerta, and H.~Peng.
\newblock Context length alone hurts llm performance despite perfect retrieval, 2025.
\newblock URL \url{https://arxiv.org/abs/2510.05381}.

\bibitem[Fang et~al.(2026)Fang, Isahagian, Jayaram, Kumar, Muthusamy, Oum, and Thomas]{fang2026trajectory}
G.~Fang, V.~Isahagian, K.~R. Jayaram, R.~Kumar, V.~Muthusamy, P.~Oum, and G.~Thomas.
\newblock Trajectory-informed memory generation for self-improving agent systems, 2026.
\newblock URL \url{https://arxiv.org/abs/2603.10600}.

\bibitem[Feng et~al.(2026)Feng, Yang, Chen, Cheng, Xu, Wan, Yan, and An]{feng2026agentocr}
L.~Feng, F.~Yang, F.~Chen, X.~Cheng, H.~Xu, Z.~Wan, M.~Yan, and B.~An.
\newblock Agentocr: Reimagining agent history via optical self-compression, 2026.
\newblock URL \url{https://arxiv.org/abs/2601.04786}.

\bibitem[Forouzandeh et~al.(2025)Forouzandeh, Peng, Moradi, Yu, and Jalili]{forouzandeh2025learning}
S.~Forouzandeh, W.~Peng, P.~Moradi, X.~Yu, and M.~Jalili.
\newblock Learning hierarchical procedural memory for llm agents through bayesian selection and contrastive refinement, 2025.
\newblock URL \url{https://arxiv.org/abs/2512.18950}.

\bibitem[Fu et~al.(2024)Fu, Kim, Kim, Sohn, Logeswaran, Bae, and Lee]{fu2024autoguide}
Y.~Fu, D.-K. Kim, J.~Kim, S.~Sohn, L.~Logeswaran, K.~Bae, and H.~Lee.
\newblock Autoguide: Automated generation and selection of context-aware guidelines for large language model agents, 2024.
\newblock URL \url{https://arxiv.org/abs/2403.08978}.

\bibitem[He et~al.(2024)He, Rungta, Koleczek, Sekhon, Wang, and Hasan]{he2024does}
J.~He, M.~Rungta, D.~Koleczek, A.~Sekhon, F.~X. Wang, and S.~Hasan.
\newblock Does prompt formatting have any impact on llm performance?, 2024.
\newblock URL \url{https://arxiv.org/abs/2411.10541}.

\bibitem[Hong et~al.(2025)Hong, Troynikov, and Huber]{hong2025context}
K.~Hong, A.~Troynikov, and J.~Huber.
\newblock Context rot: How increasing input tokens impacts {LLM} performance.
\newblock Chroma Technical Report, 2025.
\newblock URL \url{https://www.trychroma.com/research/context-rot}.

\bibitem[Hu et~al.(2025{\natexlab{a}})Hu, Van~Durme, Andreas, and Jhamtani]{hu2025sample}
M.~Y. Hu, B.~Van~Durme, J.~Andreas, and H.~Jhamtani.
\newblock Sample-efficient online learning in lm agents via hindsight trajectory rewriting, 2025{\natexlab{a}}.
\newblock URL \url{https://arxiv.org/abs/2510.10304}.

\bibitem[Hu et~al.(2025{\natexlab{b}})Hu, Liu, Yue, Zhang, Liu, Zhu, Lin, Guo, Dou, Xi, Jin, Tan, Yin, Liu, Zhang, Sun, Zhu, Sun, Peng, Cheng, Fan, Guo, Yu, Zhou, Hu, Huo, Wang, Niu, Wang, Yin, Hu, Liao, Li, Wang, Zhou, Liu, Cheng, Zhang, Gui, Pan, Zhang, Torr, Dou, Wen, Huang, Jiang, and Yan]{hu2025memory}
Y.~Hu, S.~Liu, Y.~Yue, G.~Zhang, B.~Liu, F.~Zhu, J.~Lin, H.~Guo, S.~Dou, Z.~Xi, S.~Jin, J.~Tan, Y.~Yin, J.~Liu, Z.~Zhang, Z.~Sun, Y.~Zhu, H.~Sun, B.~Peng, Z.~Cheng, X.~Fan, J.~Guo, X.~Yu, Z.~Zhou, Z.~Hu, J.~Huo, J.~Wang, Y.~Niu, Y.~Wang, Z.~Yin, X.~Hu, Y.~Liao, Q.~Li, K.~Wang, W.~Zhou, Y.~Liu, D.~Cheng, Q.~Zhang, T.~Gui, S.~Pan, Y.~Zhang, P.~Torr, Z.~Dou, J.-R. Wen, X.~Huang, Y.-G. Jiang, and S.~Yan.
\newblock Memory in the age of ai agents, 2025{\natexlab{b}}.
\newblock URL \url{https://arxiv.org/abs/2512.13564}.

\bibitem[Hutter and Pradel(2026)]{hutter2026agentstepper}
R.~Hutter and M.~Pradel.
\newblock Agentstepper: Interactive debugging of software development agents, 2026.
\newblock URL \url{https://arxiv.org/abs/2602.06593}.

\bibitem[Jiang et~al.(2023)Jiang, Wu, Lin, Yang, and Qiu]{jiang2023llmlingua}
H.~Jiang, Q.~Wu, C.-Y. Lin, Y.~Yang, and L.~Qiu.
\newblock Llmlingua: Compressing prompts for accelerated inference of large language models, 2023.
\newblock URL \url{https://arxiv.org/abs/2310.05736}.

\bibitem[Jimenez et~al.(2023)Jimenez, Yang, Wettig, Yao, Pei, Press, and Narasimhan]{jimenez2023swe}
C.~E. Jimenez, J.~Yang, A.~Wettig, S.~Yao, K.~Pei, O.~Press, and K.~Narasimhan.
\newblock Swe-bench: Can language models resolve real-world github issues?, 2023.
\newblock URL \url{https://arxiv.org/abs/2310.06770}.

\bibitem[Kang et~al.(2025)Kang, Chen, Han, Inan, Wutschitz, Chen, Sim, and Rajmohan]{kang2025acon}
M.~Kang, W.-N. Chen, D.~Han, H.~A. Inan, L.~Wutschitz, Y.~Chen, R.~Sim, and S.~Rajmohan.
\newblock Acon: Optimizing context compression for long-horizon llm agents, 2025.
\newblock URL \url{https://arxiv.org/abs/2510.00615}.

\bibitem[Khattab et~al.(2023)Khattab, Singhvi, Maheshwari, Zhang, Santhanam, Vardhamanan, Haq, Sharma, Joshi, Moazam, Miller, Zaharia, and Potts]{khattab2023dspy}
O.~Khattab, A.~Singhvi, P.~Maheshwari, Z.~Zhang, K.~Santhanam, S.~Vardhamanan, S.~Haq, A.~Sharma, T.~T. Joshi, H.~Moazam, H.~Miller, M.~Zaharia, and C.~Potts.
\newblock Dspy: Compiling declarative language model calls into self-improving pipelines, 2023.
\newblock URL \url{https://arxiv.org/abs/2310.03714}.

\bibitem[Labate et~al.(2025)Labate, de~Sousa, Fiorini, Azevedo, Thiago, and da~Silva]{labate2025solving}
A.~B. Labate, V.~M. de~Sousa, S.~R. Fiorini, L.~G. Azevedo, R.~M. Thiago, and V.~T. da~Silva.
\newblock Solving context window overflow in ai agents, 2025.
\newblock URL \url{https://arxiv.org/abs/2511.22729}.

\bibitem[{langchain-ai}(2023)]{langchainai2023langsmith}
{langchain-ai}.
\newblock langsmith-sdk.
\newblock \url{https://github.com/langchain-ai/langsmith-sdk}, 2023.
\newblock URL \url{https://github.com/langchain-ai/langsmith-sdk}.
\newblock LangSmith Client SDK Implementations.

\bibitem[{langfuse}(2023)]{langfuse2023langfuse}
{langfuse}.
\newblock langfuse.
\newblock \url{https://github.com/langfuse/langfuse}, 2023.
\newblock URL \url{https://github.com/langfuse/langfuse}.
\newblock Open source LLM engineering platform: LLM Observability, metrics, evals, prompt management, playground, datasets. Integrates with OpenTelemetry, Langchain, OpenAI SDK, LiteLLM, and more. YC W23.

\bibitem[Lindenbauer et~al.(2025)Lindenbauer, Slinko, Felder, Bogomolov, and Zharov]{lindenbauer2025complexity}
T.~Lindenbauer, I.~Slinko, L.~Felder, E.~Bogomolov, and Y.~Zharov.
\newblock The complexity trap: Simple observation masking is as efficient as llm summarization for agent context management, 2025.
\newblock URL \url{https://arxiv.org/abs/2508.21433}.

\bibitem[Liu et~al.(2023{\natexlab{a}})Liu, Lin, Hewitt, Paranjape, Bevilacqua, Petroni, and Liang]{liu2023lost}
N.~F. Liu, K.~Lin, J.~Hewitt, A.~Paranjape, M.~Bevilacqua, F.~Petroni, and P.~Liang.
\newblock Lost in the middle: How language models use long contexts, 2023{\natexlab{a}}.
\newblock URL \url{https://arxiv.org/abs/2307.03172}.

\bibitem[Liu et~al.(2023{\natexlab{b}})Liu, Yu, Zhang, Xu, Lei, Lai, Gu, Ding, Men, Yang, Zhang, Deng, Zeng, Du, Zhang, Shen, Zhang, Su, Sun, Huang, Dong, and Tang]{liu2023agentbench}
X.~Liu, H.~Yu, H.~Zhang, Y.~Xu, X.~Lei, H.~Lai, Y.~Gu, H.~Ding, K.~Men, K.~Yang, S.~Zhang, X.~Deng, A.~Zeng, Z.~Du, C.~Zhang, S.~Shen, T.~Zhang, Y.~Su, H.~Sun, M.~Huang, Y.~Dong, and J.~Tang.
\newblock Agentbench: Evaluating llms as agents, 2023{\natexlab{b}}.
\newblock URL \url{https://arxiv.org/abs/2308.03688}.

\bibitem[Liu et~al.(2025)Liu, Xu, Zhang, Chen, Feng, Chen, Guo, Yang, and Cheng]{liu2025beyond}
Y.~Liu, J.~Xu, L.~L. Zhang, Q.~Chen, X.~Feng, Y.~Chen, Z.~Guo, Y.~Yang, and P.~Cheng.
\newblock Beyond prompt content: Enhancing llm performance via content-format integrated prompt optimization, 2025.
\newblock URL \url{https://arxiv.org/abs/2502.04295}.

\bibitem[Ma et~al.(2025)Ma, Zhang, Yang, Kang, Lin, Rajmohan, and Zhang]{ma2025dover}
M.~Ma, J.~Zhang, F.~Yang, Y.~Kang, Q.~Lin, S.~Rajmohan, and D.~Zhang.
\newblock Dover: Intervention-driven auto debugging for llm multi-agent systems, 2025.
\newblock URL \url{https://arxiv.org/abs/2512.06749}.

\bibitem[Mei et~al.(2025)Mei, Yao, Ge, Wang, Bi, Cai, Liu, Li, Li, Zhang, Zhou, Mao, Xia, Guo, and Liu]{mei2025survey}
L.~Mei, J.~Yao, Y.~Ge, Y.~Wang, B.~Bi, Y.~Cai, J.~Liu, M.~Li, Z.-Z. Li, D.~Zhang, C.~Zhou, J.~Mao, T.~Xia, J.~Guo, and S.~Liu.
\newblock A survey of context engineering for large language models, 2025.
\newblock URL \url{https://arxiv.org/abs/2507.13334}.

\bibitem[{openai}(2025)]{openai2025codex}
{openai}.
\newblock codex.
\newblock \url{https://github.com/openai/codex}, 2025.
\newblock URL \url{https://github.com/openai/codex}.
\newblock Lightweight coding agent that runs in your terminal.

\bibitem[Packer et~al.(2023)Packer, Wooders, Lin, Fang, Patil, Stoica, and Gonzalez]{packer2023memgpt}
C.~Packer, S.~Wooders, K.~Lin, V.~Fang, S.~G. Patil, I.~Stoica, and J.~E. Gonzalez.
\newblock Memgpt: Towards llms as operating systems, 2023.
\newblock URL \url{https://arxiv.org/abs/2310.08560}.

\bibitem[Park et~al.(2023)Park, O'Brien, Cai, Morris, Liang, and Bernstein]{park2023generative}
J.~S. Park, J.~C. O'Brien, C.~J. Cai, M.~R. Morris, P.~Liang, and M.~S. Bernstein.
\newblock Generative agents: Interactive simulacra of human behavior, 2023.
\newblock URL \url{https://arxiv.org/abs/2304.03442}.

\bibitem[Sarthi et~al.(2024)Sarthi, Abdullah, Tuli, Khanna, Goldie, and Manning]{sarthi2024raptor}
P.~Sarthi, S.~Abdullah, A.~Tuli, S.~Khanna, A.~Goldie, and C.~D. Manning.
\newblock Raptor: Recursive abstractive processing for tree-organized retrieval, 2024.
\newblock URL \url{https://arxiv.org/abs/2401.18059}.

\bibitem[Schick et~al.(2023)Schick, Dwivedi-Yu, Dess{\`{i}}, Raileanu, Lomeli, Zettlemoyer, Cancedda, and Scialom]{schick2023toolformer}
T.~Schick, J.~Dwivedi-Yu, R.~Dess{\`{i}}, R.~Raileanu, M.~Lomeli, L.~Zettlemoyer, N.~Cancedda, and T.~Scialom.
\newblock Toolformer: Language models can teach themselves to use tools, 2023.
\newblock URL \url{https://arxiv.org/abs/2302.04761}.

\bibitem[She(2026)]{she2026agentrm}
J.~She.
\newblock Agentrm: An os-inspired resource manager for llm agent systems, 2026.
\newblock URL \url{https://arxiv.org/abs/2603.13110}.

\bibitem[Shinn et~al.(2023)Shinn, Cassano, Berman, Gopinath, Narasimhan, and Yao]{shinn2023reflexion}
N.~Shinn, F.~Cassano, E.~Berman, A.~Gopinath, K.~Narasimhan, and S.~Yao.
\newblock Reflexion: Language agents with verbal reinforcement learning, 2023.
\newblock URL \url{https://arxiv.org/abs/2303.11366}.

\bibitem[Song et~al.(2024)Song, Yin, Yue, Huang, Li, and Lin]{song2024trial}
Y.~Song, D.~Yin, X.~Yue, J.~Huang, S.~Li, and B.~Y. Lin.
\newblock Trial and error: Exploration-based trajectory optimization for llm agents, 2024.
\newblock URL \url{https://arxiv.org/abs/2403.02502}.

\bibitem[Sui et~al.(2023)Sui, Zhou, Zhou, Han, and Zhang]{sui2023table}
Y.~Sui, M.~Zhou, M.~Zhou, S.~Han, and D.~Zhang.
\newblock Table meets llm: Can large language models understand structured table data? a benchmark and empirical study, 2023.
\newblock URL \url{https://arxiv.org/abs/2305.13062}.

\bibitem[Sumers et~al.(2023)Sumers, Yao, Narasimhan, and Griffiths]{sumers2023cognitive}
T.~R. Sumers, S.~Yao, K.~Narasimhan, and T.~L. Griffiths.
\newblock Cognitive architectures for language agents, 2023.
\newblock URL \url{https://arxiv.org/abs/2309.02427}.

\bibitem[Sun et~al.(2025)Sun, Lu, Ling, Liu, Yao, Yang, and Chen]{sun2025scaling}
W.~Sun, M.~Lu, Z.~Ling, K.~Liu, X.~Yao, Y.~Yang, and J.~Chen.
\newblock Scaling long-horizon llm agent via context-folding, 2025.
\newblock URL \url{https://arxiv.org/abs/2510.11967}.

\bibitem[Suzgun et~al.(2025)Suzgun, Yuksekgonul, Bianchi, Jurafsky, and Zou]{suzgun2025dynamic}
M.~Suzgun, M.~Yuksekgonul, F.~Bianchi, D.~Jurafsky, and J.~Zou.
\newblock Dynamic cheatsheet: Test-time learning with adaptive memory, 2025.
\newblock URL \url{https://arxiv.org/abs/2504.07952}.

\bibitem[Tam et~al.(2024)Tam, Wu, Tsai, Lin, Lee, and Chen]{tam2024let}
Z.~R. Tam, C.-K. Wu, Y.-L. Tsai, C.-Y. Lin, H.-y. Lee, and Y.-N. Chen.
\newblock Let me speak freely? a study on the impact of format restrictions on performance of large language models, 2024.
\newblock URL \url{https://arxiv.org/abs/2408.02442}.

\bibitem[Trivedi et~al.(2024)Trivedi, Khot, Hartmann, Manku, Dong, Li, Gupta, Sabharwal, and Balasubramanian]{trivedi2024appworld}
H.~Trivedi, T.~Khot, M.~Hartmann, R.~Manku, V.~Dong, E.~Li, S.~Gupta, A.~Sabharwal, and N.~Balasubramanian.
\newblock Appworld: A controllable world of apps and people for benchmarking interactive coding agents, 2024.
\newblock URL \url{https://arxiv.org/abs/2407.18901}.

\bibitem[Verma(2026)]{verma2026active}
N.~Verma.
\newblock Active context compression: Autonomous memory management in llm agents, 2026.
\newblock URL \url{https://arxiv.org/abs/2601.07190}.

\bibitem[Wang et~al.(2024{\natexlab{a}})Wang, Li, Song, Xu, Tang, Zhuge, Pan, Song, Li, Singh, Tran, Li, Ma, Zheng, Qian, Shao, Muennighoff, Zhang, Hui, Lin, Brennan, Peng, Ji, and Neubig]{wang2024openhands}
X.~Wang, B.~Li, Y.~Song, F.~F. Xu, X.~Tang, M.~Zhuge, J.~Pan, Y.~Song, B.~Li, J.~Singh, H.~H. Tran, F.~Li, R.~Ma, M.~Zheng, B.~Qian, Y.~Shao, N.~Muennighoff, Y.~Zhang, B.~Hui, J.~Lin, R.~Brennan, H.~Peng, H.~Ji, and G.~Neubig.
\newblock Openhands: An open platform for ai software developers as generalist agents, 2024{\natexlab{a}}.
\newblock URL \url{https://arxiv.org/abs/2407.16741}.

\bibitem[Wang et~al.(2024{\natexlab{b}})Wang, Mao, Fried, and Neubig]{wang2024agent}
Z.~Z. Wang, J.~Mao, D.~Fried, and G.~Neubig.
\newblock Agent workflow memory, 2024{\natexlab{b}}.
\newblock URL \url{https://arxiv.org/abs/2409.07429}.

\bibitem[Wei et~al.(2022)Wei, Wang, Schuurmans, Bosma, Ichter, Xia, Chi, Le, and Zhou]{wei2022chain}
J.~Wei, X.~Wang, D.~Schuurmans, M.~Bosma, B.~Ichter, F.~Xia, E.~Chi, Q.~Le, and D.~Zhou.
\newblock Chain-of-thought prompting elicits reasoning in large language models, 2022.
\newblock URL \url{https://arxiv.org/abs/2201.11903}.

\bibitem[Wu et~al.(2023)Wu, Bansal, Zhang, Wu, Li, Zhu, Jiang, Zhang, Zhang, Liu, Awadallah, White, Burger, and Wang]{wu2023autogen}
Q.~Wu, G.~Bansal, J.~Zhang, Y.~Wu, B.~Li, E.~Zhu, L.~Jiang, X.~Zhang, S.~Zhang, J.~Liu, A.~H. Awadallah, R.~W. White, D.~Burger, and C.~Wang.
\newblock Autogen: Enabling next-gen llm applications via multi-agent conversation, 2023.
\newblock URL \url{https://arxiv.org/abs/2308.08155}.

\bibitem[Xiao et~al.(2025)Xiao, Gao, Peng, and Xiong]{xiao2025reducing}
Y.-A. Xiao, P.~Gao, C.~Peng, and Y.~Xiong.
\newblock Reducing cost of llm agents with trajectory reduction, 2025.
\newblock URL \url{https://arxiv.org/abs/2509.23586}.

\bibitem[Xu et~al.(2025)Xu, Liang, Mei, Gao, Tan, and Zhang]{xu2025mem}
W.~Xu, Z.~Liang, K.~Mei, H.~Gao, J.~Tan, and Y.~Zhang.
\newblock A-mem: Agentic memory for llm agents, 2025.
\newblock URL \url{https://arxiv.org/abs/2502.12110}.

\bibitem[Yang et~al.(2024)Yang, Jimenez, Wettig, Lieret, Yao, Narasimhan, and Press]{yang2024swe}
J.~Yang, C.~E. Jimenez, A.~Wettig, K.~Lieret, S.~Yao, K.~Narasimhan, and O.~Press.
\newblock Swe-agent: Agent-computer interfaces enable automated software engineering, 2024.
\newblock URL \url{https://arxiv.org/abs/2405.15793}.

\bibitem[Yang et~al.(2026)Yang, Wang, Zhang, Yu, Wu, Gui, Yang, Cen, Feng, Wen, Wang, Zhong, Ren, Zhang, and Tang]{yang2026tracesir}
S.-X. Yang, C.~Wang, H.~Zhang, W.~Yu, L.~Wu, J.~Gui, D.~Yang, Y.~Cen, Z.~Feng, B.~Wen, Y.~Wang, L.~Zhong, J.~Ren, L.~Zhang, and J.~Tang.
\newblock Tracesir: A multi-agent framework for structured analysis and reporting of agentic execution traces, 2026.
\newblock URL \url{https://arxiv.org/abs/2603.00623}.

\bibitem[Yao et~al.(2022)Yao, Zhao, Yu, Du, Shafran, Narasimhan, and Cao]{yao2022react}
S.~Yao, J.~Zhao, D.~Yu, N.~Du, I.~Shafran, K.~Narasimhan, and Y.~Cao.
\newblock React: Synergizing reasoning and acting in language models, 2022.
\newblock URL \url{https://arxiv.org/abs/2210.03629}.

\bibitem[Yao et~al.(2024)Yao, Shinn, Razavi, and Narasimhan]{yao2024tau}
S.~Yao, N.~Shinn, P.~Razavi, and K.~Narasimhan.
\newblock $\tau$-bench: A benchmark for tool-agent-user interaction in real-world domains, 2024.
\newblock URL \url{https://arxiv.org/abs/2406.12045}.

\bibitem[Ye et~al.(2025)Ye, Zhang, Li, Yin, Tao, Zhao, Su, Zhang, Qiao, Wang, Xie, Huang, Chen, Zhou, and Jiang]{ye2025agentfold}
R.~Ye, Z.~Zhang, K.~Li, H.~Yin, Z.~Tao, Y.~Zhao, L.~Su, L.~Zhang, Z.~Qiao, X.~Wang, P.~Xie, F.~Huang, S.~Chen, J.~Zhou, and Y.~Jiang.
\newblock Agentfold: Long-horizon web agents with proactive context management, 2025.
\newblock URL \url{https://arxiv.org/abs/2510.24699}.

\bibitem[Yuan et~al.(2026)Yuan, Yuan, and Xie]{yuan2026rpms}
Z.~Yuan, S.~Yuan, and L.~Xie.
\newblock Rpms: Enhancing llm-based embodied planning through rule-augmented memory synergy, 2026.
\newblock URL \url{https://arxiv.org/abs/2603.17831}.

\bibitem[Zeng et~al.(2026)Zeng, Huang, and He]{zeng2026loca}
W.~Zeng, Y.~Huang, and J.~He.
\newblock Loca-bench: Benchmarking language agents under controllable and extreme context growth, 2026.
\newblock URL \url{https://arxiv.org/abs/2602.07962}.

\bibitem[Zhang et~al.(2025{\natexlab{a}})Zhang, Wang, Chen, Zhou, Wang, and Yan]{zhang2025agentracer}
G.~Zhang, J.~Wang, J.~Chen, W.~Zhou, K.~Wang, and S.~Yan.
\newblock Agentracer: Who is inducing failure in the llm agentic systems?, 2025{\natexlab{a}}.
\newblock URL \url{https://arxiv.org/abs/2509.03312}.

\bibitem[Zhang et~al.(2025{\natexlab{b}})Zhang, Hu, Upasani, Ma, Hong, Kamanuru, Rainton, Wu, Ji, Li, Thakker, Zou, and Olukotun]{zhang2025agentic}
Q.~Zhang, C.~Hu, S.~Upasani, B.~Ma, F.~Hong, V.~Kamanuru, J.~Rainton, C.~Wu, M.~Ji, H.~Li, U.~Thakker, J.~Zou, and K.~Olukotun.
\newblock Agentic context engineering: Evolving contexts for self-improving language models, 2025{\natexlab{b}}.
\newblock URL \url{https://arxiv.org/abs/2510.04618}.

\bibitem[Zhao et~al.(2023)Zhao, Huang, Xu, Lin, Liu, and Huang]{zhao2023expel}
A.~Zhao, D.~Huang, Q.~Xu, M.~Lin, Y.-J. Liu, and G.~Huang.
\newblock Expel: Llm agents are experiential learners, 2023.
\newblock URL \url{https://arxiv.org/abs/2308.10144}.

\bibitem[Zheng et~al.(2023)Zheng, Yin, Xie, Sun, Huang, Yu, Cao, Kozyrakis, Stoica, Gonzalez, Barrett, and Sheng]{zheng2023sglang}
L.~Zheng, L.~Yin, Z.~Xie, C.~Sun, J.~Huang, C.~H. Yu, S.~Cao, C.~Kozyrakis, I.~Stoica, J.~E. Gonzalez, C.~Barrett, and Y.~Sheng.
\newblock Sglang: Efficient execution of structured language model programs, 2023.
\newblock URL \url{https://arxiv.org/abs/2312.07104}.

\bibitem[Zhou et~al.(2023)Zhou, Xu, Zhu, Zhou, Lo, Sridhar, Cheng, Ou, Bisk, Fried, Alon, and Neubig]{zhou2023webarena}
S.~Zhou, F.~F. Xu, H.~Zhu, X.~Zhou, R.~Lo, A.~Sridhar, X.~Cheng, T.~Ou, Y.~Bisk, D.~Fried, U.~Alon, and G.~Neubig.
\newblock Webarena: A realistic web environment for building autonomous agents, 2023.
\newblock URL \url{https://arxiv.org/abs/2307.13854}.

\end{thebibliography}

\end{document}